# RSLaA: Self-Reconfiguring Roadside LiDARs as Agents for 3D Traffic Perception

Yang Ma[1*] Yichen Sun[1] Yubing Zheng[1] Zhongxiang Feng[1] Zeyang Cheng[1] Said M. Easa[2]
[1]Hefei University of Technology [2]Toronto Metropolitan University
{mayang, ybzheng, fzx, chengzeyang}@hfut.edu.cn {seasa}@torontomu.ca

## Abstract

*Cooperative perception with roadside LiDARs (RSLs) extends the sensing range of autonomous vehicles and resolves onboard-sensor occlusions, particularly at curves and intersections. Because infrastructure deployment is costly, existing work optimizes RSL configurations once, against road geometry and aggregated traffic statistics, and treats them as fixed thereafter. Traffic flow, however, is inherently dynamic, and a placement optimal on average can be far from optimal at any given moment. We introduce* ***RSLaA (Roadside LiDARs as Agents)****, a framework in which each RSL adaptively adjusts its pose to the evolving traffic scene over decision-time windows. Since ground truth is unavailable at decision time, the framework centers on surrogate metrics computed solely from a probabilistic occupancy grid of historical observations, which predicts the next-window perception performance of candidate poses. We construct a co-simulation benchmark pairing five candidate surrogates with detector-agnostic ground-truth perception labels at every decision window, identifying the multi-view coverage metric $J_{mv}$ as the most reliable predictor. Upon this validated surrogate we build two adaptive configuration methods, a receding-horizon (MPC) optimizer and a lightweight MAPPO-based multi-agent reinforcement learning scheme with centralized training and decentralized execution. Across 10 sites and 89 six-hour traffic scenarios (890 evaluation runs), adaptive configuration consistently improves the composite perception score over static deployment. Our results indicate that RSLs could be benefitted from being treated as agents.* https://github.com/yangma93/RSLaA

## 1. Introduction

Reliable 3D perception of the surrounding traffic participants is a prerequisite for safe autonomous driving. However, onboard perception is fundamentally constrained by its ego-centric viewpoint: sensing range is limited, and occlusion caused by other road users, buildings, or road curvature can leave safety-critical regions unobserved. To overcome these limitations, cooperative vehicle-infrastructure perception has attracted increasing attention, in which roadside sensors, particularly LiDARs mounted on poles, gantries, or signal masts, provide an elevated, fixed-coordinate vantage point that complements the onboard view [1][2][3][4]. By fusing point clouds from multiple roadside LiDARs (RSL(s)), an infrastructure-assisted system can see through occlusions and cover regions that are difficult for any single sensor, which is especially valuable at curves, intersections, and other occlusion-dense areas.

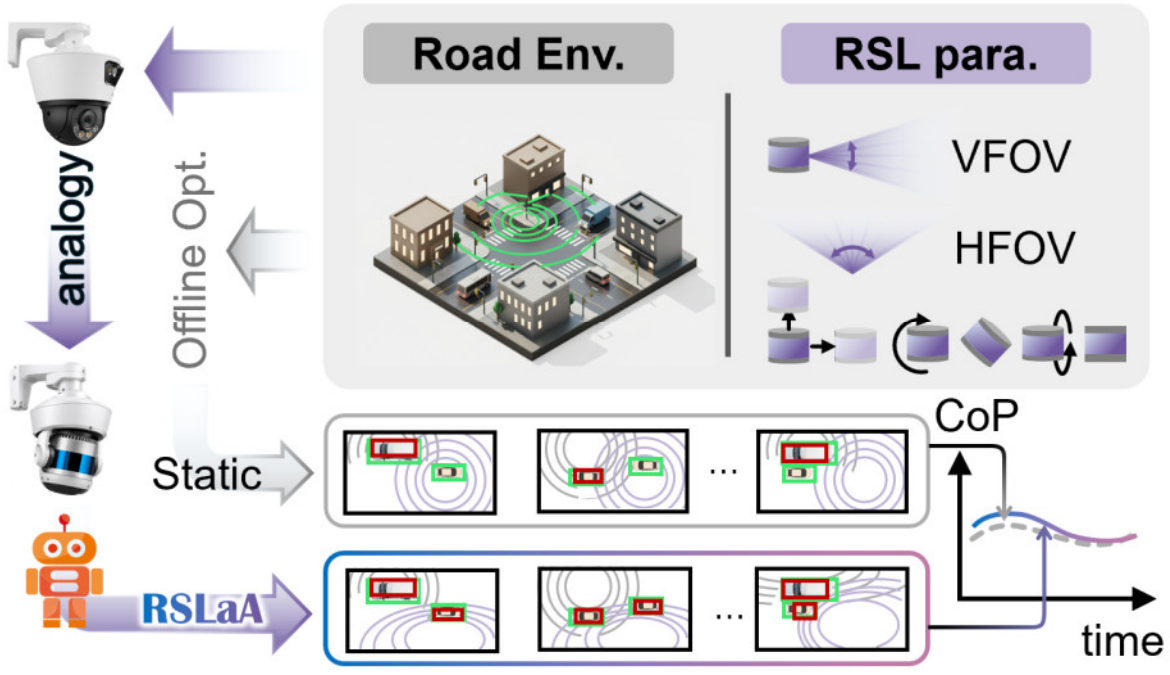

**Figure 1.** Concept of Roadside LiDARs as Agents.

Because RSL systems are expensive to deploy and maintain, a growing body of work has studied how to configure a limited number of RSLs to maximize their cooperative sensing capability [5][6][7][8][9][10]. These studies formulate sensor configuration as a *static* placement optimization problem: given the road geometry, a coverage or occlusion model, and aggregated traffic statistics, a fixed installation position and orientation is solved once by maximizing point cloud coverage, optimizing detection-oriented surrogate metrics, or searching configurations with evolutionary or learning-based solvers. The position and orientation will remain unchanged after deployment.

We argue that this static paradigm leaves a substantial fraction of the cooperative potential of roadside LiDARs unexploited. Traffic flow is inherently dynamic, traffic volume, speed, and composition vary across time, and the occlusion relationships between vehicles, roadside obstacles, and sensors evolve continuously as vehicles move. A placement that is optimal on average over historical statistics can therefore be far from optimal at any given moment, e.g., when a dense traffic stream systematically shadows a critical region from a fixed viewpoint.

We may transfer ideas in other dominant roadside sensors to RSL-based cooperative perception. For instance, pan-tilt-zoom cameras routinely adjust their poses to track targets of interest to address the problem of limited FOV. Meanwhile, two practical trends make pose-adaptive RSL increasingly viable: (i) multi-degree-of-freedom adjustable mounts are becoming affordable relative to the sensor cost, and (ii) the rapid growth of real and synthetic co-simulation data enables training and validating perception and decision models for such systems. These observations motivate **a new question**: *can roadside LiDARs, like pan-tilt-zoom cameras, adapt their poses to the evolving traffic scene so that the cooperative system continuously operates near its instantaneous optimum?*

Answering this question faces a fundamental obstacle that distinguishes it from static placement optimization: *there is no ground truth (GT) at decision time*. Offline placement studies can evaluate a configuration by running full perception pipelines against annotated data [7][8][9], but a deployed adaptive system cannot know whether a candidate pose will improve perception owing to the absence of GT BBoxes. The system must therefore assess itself through a surrogate metric (SM): a quantity computable solely from historical, causally available information, which predicts the perception performance associated with candidate poses in the next decision window. Since storing raw point cloud streams is prohibitive, we compress the historical observations into a probabilistic occupancy grid (POG) accumulated from the LiDAR point cloud stream, and define SM over this compact history representation. This predictive, history-only requirement also differentiates our surrogate from existing placement metrics [8][9], which are evaluated offline and need not guarantee temporal causality.

An orthogonal alternative is to learn a predictive *world model* of the traffic scene and plan RSL configurations against simulated rollouts; we deliberately refrain from this direction, because training and hosting such models at every roadside node would substantially raise the already significant infrastructure-side deployment and operation cost, the very expenditure that RSL placement optimization seeks to minimize.

In this paper, we introduce **Roadside LiDARs as Agents (RSLaA)**, as illustrated in Figure 1, a framework that formulates adaptive RSL pose configuration as a sequential decision-making problem and provides a complete pipeline from problem definition to empirical validation. Our contributions are summarized as follows:

- We define the *RSLaA* problem, treating each RSL as an agent that adapts its pose to dynamic traffic, and decompose it into a two-stage process of surrogate-metric computation and adaptive configuration, highlighting the necessity of a causal, history-only self-evaluation signal.
- We construct a co-simulation benchmark for surrogate metrics of adaptive RSL configuration, in which historical point cloud information is represented as a probabilistic occupancy grid; candidate surrogates are paired with detector-agnostic GT perception labels at every decision window, enabling correlation-based selection of the decision core.
- Building on the selected surrogate, we develop model predictive control (MPC) and multi-agent reinforcement learning (MARL) adaptive configuration frameworks, and conduct comparative experiments over 10 sites and hundreds of scenario configurations, together with a SHAP-based analysis of the factors governing the adaptive gain.

# 2. Related Work

## 2.1. LiDAR-based Cooperative Perception

Cooperative perception (CoP) aggregates sensing information from multiple connected agents to overcome the occlusion and range limits of single-vehicle perception. The field has been advanced by a series of datasets: simulated benchmarks such as OPV2V [11] established standard fusion pipelines and evaluation protocols for vehicle-to-vehicle collaboration, while real-world datasets including DAIR-V2X [1], V2V4Real [4], V2X-Seq [12], TUMTraf V2X [3], and CoInfra [13] have extended the setting to vehicle-infrastructure cooperation with calibrated roadside and onboard sensors under realistic traffic. Methodologically, existing work focuses on how to fuse given observations: early fusion of raw point clouds, late fusion of per-agent detections, and intermediate fusion of learned features, the latter exemplified by attentive fusion in OPV2V [11], the vision-transformer architecture of V2X-ViT [2], the spatial-confidence-guided communication of Where2comm [14], cooperative BEV segmentation with CoBEVT [15], and temporal feature fusion for vehicle-infrastructure links [16]. A common premise of all these works is that the data acquisition geometry, that is where the sensors are and where they look, is fixed, and performance gains are sought purely at the algorithmic level. Our work is complementary. We treat the acquisition geometry itself as a decision variable and ask how roadside LiDARs should actively adjust it in response to dynamic traffic.

## 2.2. Static Optimization of Roadside Sensor Placement

Recognizing that roadside perception quality depends strongly on where sensors are installed, a substantial literature studies the one-shot optimization of roadside sensor placement. Coverage-driven approaches model the sensing footprint of cameras, radars, and LiDARs and solve for placements that maximize covered road area under budget constraints [5], or incorporate explicit three-dimensional occlusion models coupled with microscopic traffic simulation to quantify and optimize detection and tracking performance [6]; integer programming formulations jointly decide sensor locations, heights, and types under cost constraints [17], and Pareto-optimal placement families have been explored with evolutionary algorithms [19]. A second line of work evaluates placements by their downstream perception utility: realistic LiDAR simulation libraries enable detection-based comparison of candidate infrastructure placements [7], while lightweight surrogate metrics, such as the information-theoretic entropy measure of [8] and the semantic-occupancy-based metric of Place3D [9], predict the detection performance of a configuration without running full perception pipelines; recent work further refines this idea by optimizing the spatial distribution of points on vehicle surfaces directly [10]. Learning-based solvers have also been proposed for on-vehicle sensor placement [18]. Despite their diversity, these approaches share a static formulation: the objective is built from road geometry and aggregated traffic statistics, the configuration is solved once, and no mechanism exists to adapt the sensors after deployment. As argued in Sec. 1, this assumption is increasingly at odds with the dynamic nature of traffic flow and occlusion, and recent reviews of roadside LiDAR perception likewise identify deployment optimization under dynamic conditions as an open challenge [20].

## 2.3. Adaptive Sensor Configuration

Adjusting a sensor to acquire more informative observations has a long history in computer vision and robotics. In active object recognition and reconstruction, the next-best-view problem seeks the viewpoint that maximizes information gain for the current task, addressed by classic greedy planning over occupancy models [21], appearance-driven one-shot view selection [22], supervised learning of view policies [23], and deep reinforcement learning [24]. Perception-aware control embeds sensing objectives into motion optimization: model predictive control has been used to keep targets visible during quadrotor flight [25] and to compensate latency in active visual tracking [26]. In the sensor-management literature, agile sensors are dynamically retasked, repositioned, re-pointed, or re-modulated, in response to an evolving environment, with information-theoretic objectives for multi-target tracking [27] and, more recently, deep reinforcement learning solutions in continuous action spaces [28]. Multi-agent reinforcement learning provides scalable machinery for such cooperative sequential decisions, with centralized-training-decentralized-execution algorithms such as MAPPO [29] achieving strong performance across cooperative benchmarks. Our work inherits this active-perception perspective but differs in three essential aspects. First, our agents are infrastructure-mounted LiDARs with small, physically constrained pose margins rather than freely moving robots, so adaptation must be realized through incremental pose adjustments over decision windows. Second, the objective is the team-level 3D perception quality of a cooperative multi-LiDAR system over an entire traffic scene, not the information gain of a single sensor about a single object. Third, and most critically, the decision signal must be computable causally from historical observations alone, without ground truth at decision time; this requirement motivates our surrogate-metric benchmark and the MPC and MARL frameworks built upon it, which, to our knowledge, constitute the first systematic study of adaptive RSL pose configuration for dynamic traffic scenes.

# 3. Method

RSLaA decomposes the problem into two coupled stages. First, at the problem formulation level, we discretize continuous adaptation into decision windows of length $w$, and formalize the SM that maps historical POG information to a prediction of next-window and pose-dependent perception performance. Second, at the solution level, we build the decision-making machinery on top of a validated SM.

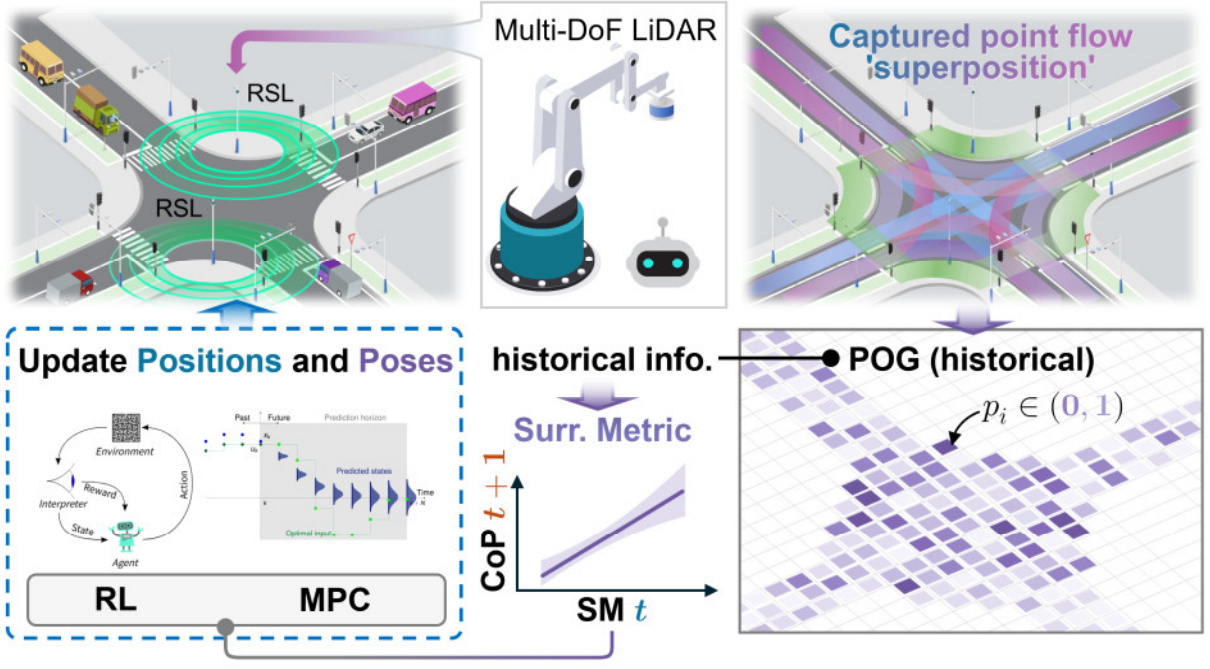

**Figure 2.** General workflow of RSLaA.

Specifically, RSLaA consists of three components. Sec. 3.1 formalizes adaptive RSL pose configuration as a sequential decision problem and identifies the necessity of a SM. Sec. 3.2 introduces the candidate SMs and the co-simulation benchmark built to validate them, with the definition of the selected metric $J_{\mathrm{mv}}$. Sec. 3.3 presents the two adaptive configuration methods built upon the validated surrogate: MPC and MARL. The general workflow of RSLaA is graphically shown in Figure 2.

## 3.1. Problem Formulation

We consider a roadside CoP system of $N$ LiDARs installed along a road section. The pose parameter of LiDAR $n$, denoted as $\mathcal{P}_n$ and $\mathcal{A}_n$, comprises its three-dimensional position and roll–pitch–yaw angles, respectively. Continuous operation is partitioned into decision windows of length $w$, and the pose of each LiDAR may be adjusted only at window boundaries. This discretization reflects three practical considerations: traffic statistics relevant to perception evolve at the minute scale rather than the frame scale; pose adjustment incurs mechanical and computational cost; and overly frequent re-aiming brings negligible benefit while complicating multi-sensor registration and fusion. Each adjustment is bounded by the physical margins of the mounting device, i.e. position increments within $s_p$ per translation axis and angular increments within $s_a$ per rotation angle, measured relative to the pose of the previous window.

After each window, the fused multi-LiDAR point clouds are accumulated into a POG, which stores for each voxel $i$ an occupancy probability $p_i$ in [0,1] together with an observation mask $o_i$ indicating whether the voxel has ever fallen within the field of view of any RSL; bird's-eye-view (BEV) column occupancy is obtained by taking the maximum probability along the height axis. The POG is the sole information substrate for decision making. POG is causally available, far more compact than raw point cloud streams, and preserves the spatial distribution of both static structure and historical traffic activity.

Let $S$ denote the perception performance achieved in a window under a given pose configuration, measured against ground truth (Sec. 3.2). The adaptive configuration problem seeks a pose sequence that maximizes the cumulative perception performance under the adjustment constraints as:

$$\max_{\{\mathcal{P}^{k+1},\mathcal{A}^{k+1}\}} \sum_{k=1}^{K} S\left(\mathcal{P}^{k+1},\mathcal{A}^{k+1}\right) \tag{1}$$

$$\text{s.t. } \mathcal{P}^{k+1} = \mathcal{P}^{k} + \Delta\mathcal{P}^{k}, \mathcal{A}^{k+1} = \mathcal{A}^{k} + \Delta\mathcal{A}^{k},\ |\Delta\text{pos}_n^k| \le s_p,\ |\Delta\text{ang}_n^k| \le s_a$$

The difficulty is that $S$ is undefined at decision time. Scoring a candidate configuration requires executing it and acquiring the next-window observations. Any operational method must therefore replace $S$ with a SM $J$ that maps the history $\mathcal{H}_k$ (the sequence of past POGs) and a candidate configuration to a prediction of its next-window performance as:

$$J(\mathcal{H}_k,\mathcal{P},\mathcal{A}) \approx S(\mathcal{P},\mathcal{A}) \tag{2}$$

where $\mathcal{P},\mathcal{A} \in \{\text{candidate configurations at window } k\}$. Such that the ranking of candidates by $J$ predicts their ranking by $S$. From the multi-agent perspective, each LiDAR is an agent whose joint action is the pose adjustment and whose shared objective is the team-level perception quality; since no agent observes the full traffic state and decisions are made simultaneously from compressed history, the problem is a decentralized partially observable Markov decision process.

## 3.2. SMs and Benchmark Construction

**Surrogate metric**. We propose and implement five SMs, each encoding a distinct hypothesis about what makes a configuration perceptually valuable, so that the benchmark can arbitrate between competing design intuitions rather than endorse a single one a priori. $J_{\text{EIG}}$ approximates the expected information gain, prioritizing coverage of voxels whose occupancy state is still uncertain. $J_{\text{cov}}$ measures task-weighted coverage, i.e., the fraction of the task region, defined by the historical trajectory corridor and the sensors' fields of view, that is actually visible. $J_{\text{mv}}$ augments coverage with multi-view complementarity, rewarding configurations in which LiDAR pairs observe the same location under favorable triangulation angles. $J_{\text{dyn}}$ concentrates visible coverage on historically dynamic regions, where traffic participants have actually been active. $J_{\text{traj}}$ evaluates the continuity of visibility along predicted traffic corridors, penalizing consecutive low-visibility segments that would become perception blind spots for approaching vehicles. The definitions, implementation details, and degenerate-handling rules of all five metrics are provided in the supplementary material. Here we give the complete definition of $J_{\text{mv}}$, which the benchmark identifies as the decision core of both adaptive methods.

The $J_{\text{mv}}$-metric measures both how well a candidate configuration covers the task region and, crucially, how favorably the $N$ LiDARs triangulate it:

$$J_{\text{mv}} = \frac{\sum_k w_k\, q_k^{\text{col}}}{\sum_k w_k} + \lambda\, \frac{\sum_{a<b}\sum_k w_k\, q_{a,k}^{\text{col}}\, q_{b,k}^{\text{col}}\, |\sin\theta_{ab,k}|}{\sum_k w_k} \tag{3}$$

$$q_k^{\text{col}} = 1 - \prod_{n=1}^{N}\left(1 - q_{n,k}^{\text{col}}\right) \tag{4}$$

where $k$ indexes the BEV columns of the task region, and $w_k$ is the column task weight, equal to 1 for columns lying in the trajectory-corridor task region and within the field of view and range of at least one LiDAR, and 0 otherwise; $q_{n,k}^{\text{col}}$ is the column-level soft visibility of column $k$ by LiDAR $n$ under the candidate pose, computed as a distance-decayed transmittance along the 2D sight line estimated over the historical column occupancy, masked by the sensor's field of view and range limits; $q_k^{\text{col}}$ is the joint column visibility under an observation-independence assumption; $\theta_{ab,k}$ is the angle between the sight lines of the LiDAR pair $(a, b)$ at column $k$; and $\lambda$ balances coverage against complementarity. The first term is a weighted coverage score, while the second term rewards configurations in which the same location is observed by two sensors at near-orthogonal angles, meaning that the geometry under which occlusion is most easily resolved and 3D localization error is minimized.

**Benchmark construction.** To establish whether any surrogate deserves to guide decisions, we construct a benchmark in which candidate surrogates are paired with ground-truth perception labels at every decision window, as illustrated in Figure 3. Illustration of benchmark creation.

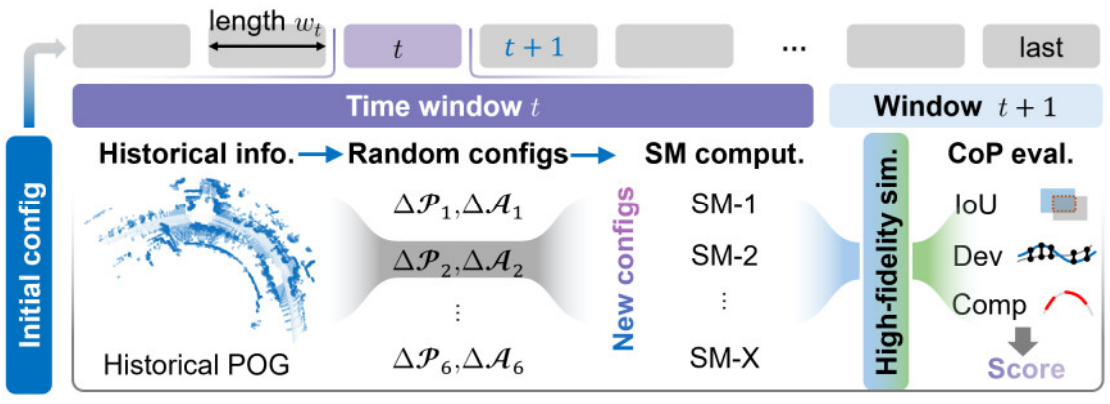

**Figure 3.** Illustration of benchmark creation

The benchmark is built through co-simulations. Specifically, microscopic traffic simulation generates vehicle trajectories for 6-hour scenarios at several real sites reconstructed from point clouds, and a high-fidelity LiDAR simulator renders the point clouds of 2 to 4 heterogeneous sensors (VLP-16, HDL-32, OS1-64, RS-128) under arbitrary poses. Each run starts from a randomized pole-mounted initial configuration. From the second window onward, six candidate adjustment plans are sampled within the pose margins around the previously selected configuration; the five surrogates are computed for each candidate using only the historical POGs, and every candidate is then executed in simulation for the next window to obtain its perception scores. To keep the labels detector-agnostic, no learned detector is used. For each vehicle we fit its BBox directly from the simulated points belonging to that target and compare the fitted box against the GT envelope, yielding per-window IoU, center deviation (Dev), and sequence completeness (Comp).

The best-scoring candidate is executed, the POG is rebuilt, and the process rolls forward until the scenario ends; the whole procedure is repeated for window lengths $w$ of 5, 15, and 30 min. This candidate-level pairing of SMs computed from history only and labels measured strictly in the future window enables direct correlation analysis. We compute Pearson and Spearman correlations between each surrogate and the perception labels, stratified by window length and adjustment margin, and additionally examine selection validity, i.e., the true-score rank of the surrogate-preferred candidate among the six candidates.

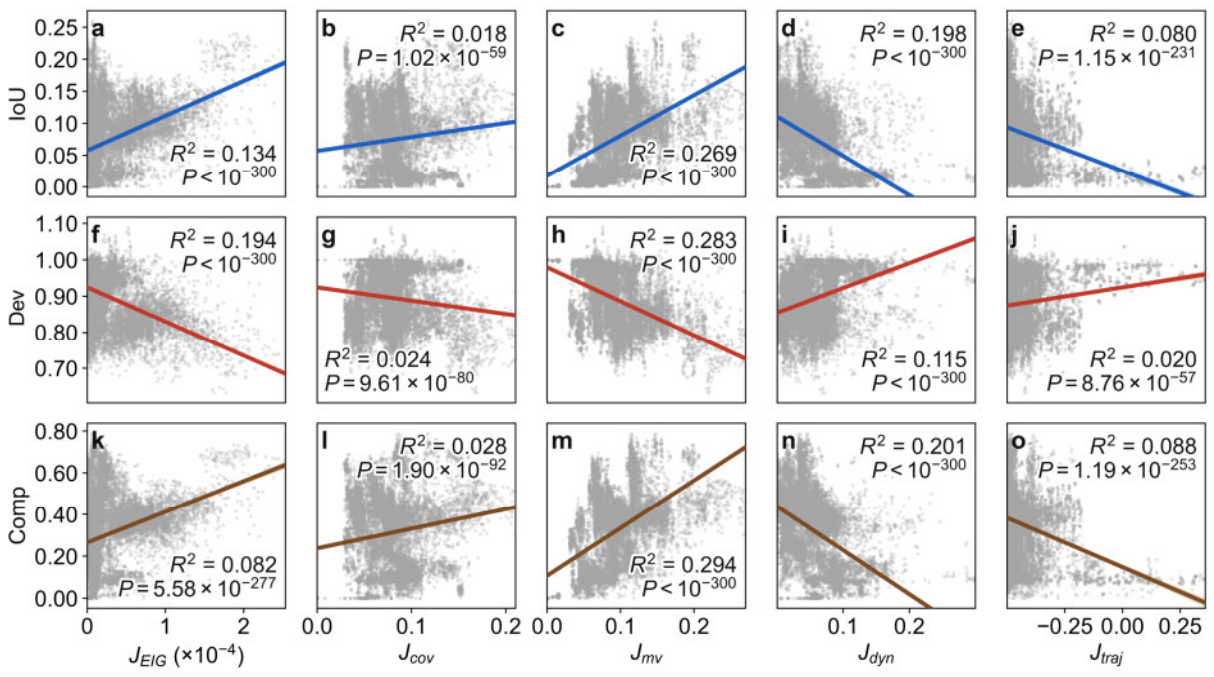

**Figure 4.** Correlation results of five SM

The benchmark covers 17 point-cloud sites and 25 traffic scenarios, totaling 2,522-time windows and 14,694 candidate LiDAR configurations, which yield 15 SM-perception metric pairs per candidate (220,410 pairs in total, 211,938 after removing degenerate surrogate values). Please refer to the supplementary material for details.

The benchmark shows that coverage-type surrogates correlate strongly with perception performance overall, while $J_{\mathrm{mv}}$ (Pearson's $r$ of 0.571, p<0.05) offers the most favorable balance between correlation strength and relevance to moving-vehicle perception (see Figure 4), and is therefore adopted as the shared optimization core of the MPC and MARL methods below.

## 3.3. Adaptive Configuration of RSL

**MPC approach**. The MPC branch instantiates Eq. (1) as a receding-horizon reconfiguration scheme. Strictly speaking, it is MPC-form parameter adaptation rather than trajectory-tracking control. The state is the pose configuration itself; the action is the pose increment, the constraints are per-dimension box constraints around the initial pose, and the role of the predictive model is played by the historical POG together with the SM, meaning that no vehicle-dynamics model or terminal cost is involved, and the prediction horizon is one window. At each window boundary $k{\geq}2$, and before the next window is simulated, the controller solves a constrained joint optimization over all $N$ LiDARs:

$$\{\Delta\mathcal{P}_n^k, \Delta\mathcal{A}_n^k\}^* = \arg\max_{\{\Delta\mathcal{P}_n, \Delta\mathcal{A}_n\}} \tilde{J}_{\mathrm{mv}}(\mathcal{H}_k, \mathcal{P}^k + \Delta\mathcal{P}, \mathcal{A}^k + \Delta\mathcal{A}) \quad (5)$$
$$\text{s.t. } |\Delta\mathrm{pos}_n| \leq s_p, \ |\Delta\mathrm{ang}_n| \leq s_a$$

where $\tilde{J}_{\mathrm{mv}}$ is a column-level approximation of $J_{\mathrm{mv}}$ to accelerate computations: visibility is evaluated at the BEV-column level within the task region with uniformly subsampled columns (at most 12,000, fixed seed) and reduced line-of-sight samples, which lowers a single evaluation to the order of ten milliseconds and makes the thousands of objective evaluations required per window affordable; the configuration finally executed is always re-scored with the exact voxel-level metric. After the new window is simulated, the POG is rebuilt and the horizon recedes. The MPC loop is model-free with respect to traffic. The historical POG plays the role of the predictive model, and requires no training, but performs myopic one-step optimization at every window.

**MARL approach.** To amortize the online optimization and to capture longer-horizon coordination, we cast the problem as cooperative MARL and adopt a centralized-training, decentralized-execution architecture based on MAPPO [28], which extends PPO [30] to cooperative multi-agent settings. The framework is shown in Figure 5. Framework of MARL method for RSLaA The design is deliberately lightweight so that per-window inference latency stays negligible relative to the window length.

*Observation.* Each agent $i$ receives a 127-dimensional observation $o_i$ comprising: (i) its 6-dimensional pose relative to a designated reference unit (i.e. LiDAR#0), with translations expressed in the reference heading frame and angular differences wrapped to $(-\pi, \pi)$; (ii) a 7×7 local BEV occupancy patch centered on its own position and rotated to the reference heading; and (iii) a 72-dimensional global encoding shared by the team, consisting of an 8×8 coarse BEV (8 m cells covering 64 m × 64 m) of the joint POG plus 8 occupancy statistics. The "relative pose-local patch-shared global encoding" structure makes the learned mapping invariant to global translation and rotation, keeps the input distribution consistent across scenes with different grid extents, and provides every agent with a common situational picture while retaining its own local detail (i.e. global consensus with local specialization), without any explicit communication channel.

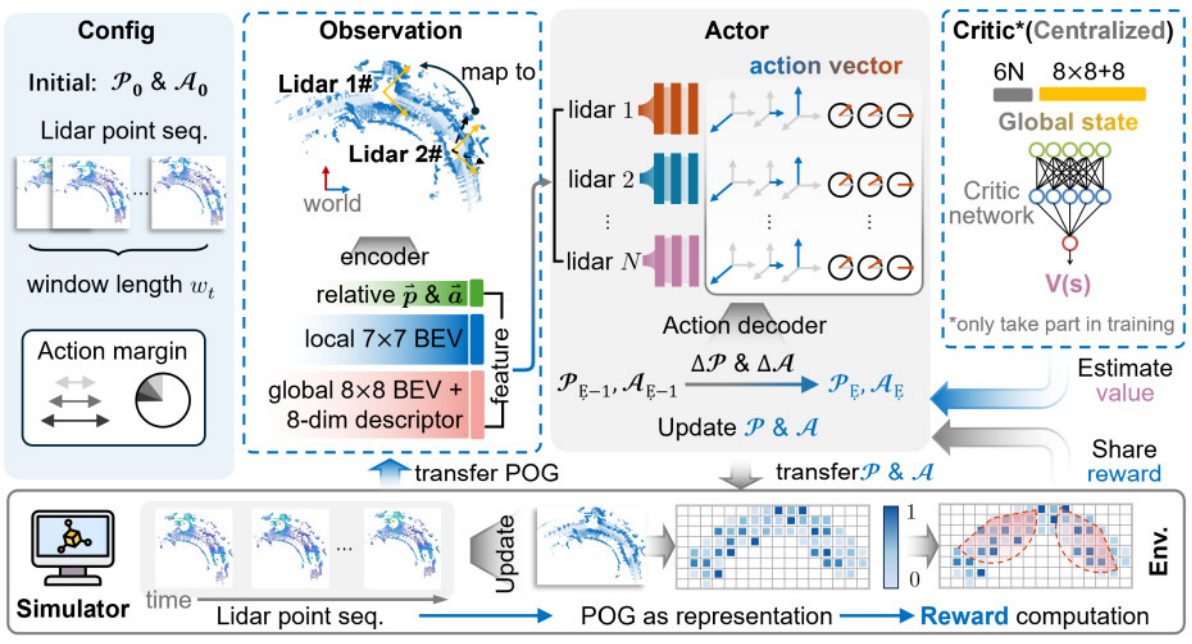

**Figure 5.** Framework of MARL method for RSLaA

*Action.* Each actor outputs a normalized 6-dimensional action $a_i$ in $[-1, 1]$, decoded into a physical pose increment relative to the previous window: $\Delta x_i = a_i[0] \cdot s_p$, $\Delta y_i = a_i[1] \cdot s_p$, $\Delta z_i = a_i[2] \cdot s_p$, $\Delta\alpha_i^{r,p,y} = a_i[3{:}6] \cdot s_a$, with the horizontal components expressed in the reference unit's local frame and rotated to the world frame by its yaw, so that the same policy produces consistent physical behavior on roads of different orientations. Defining actions as increments rather than absolute poses keeps the action space local and smooth: the policy only needs to learn "which way to adjust", not to re-output absolute coordinates.

*Reward.* The team shares a scalar reward $r_t$ equal to the $J_{\text{mv}}$ of the current window, computed from the latest POG with a short history, so maximizing the discounted return is equivalent to maximizing the cumulative $J_{\text{mv}}$. A shared reward eliminates per-agent competition and ensures that any adjustment improving team perception reinforces the contributing behavior of all actors.

*Actor and critic.* Each LiDAR has an independent actor network, a two-hidden-layer MLP (127-128-128 units, tanh activation) with a linear mean head producing the 6-dimensional action mean and a state-independent learnable log-standard-deviation, forming a diagonal Gaussian policy during training and a deterministic mean policy at deployment. Independent actors allow heterogeneous sensor models to specialize into complementary roles, e.g., long-range units favoring wide coverage and short-range units filling near-field blind regions. The two-layer width-128 architecture is sufficient for the "relative pose + local geometry→adjustment direction" mapping while avoiding overfitting under limited episodes. The centralized critic takes the global state, that is the concatenation of all agents' relative poses and the shared global encoding, $6N + 72$ dimensions, through a wider two-hidden-layer MLP (256 units per layer, tanh) to a scalar value $V(s)$. The critic serves two purposes: (i) credit assignment under the shared team reward, since conditioning on the joint configuration lets each agent's advantage reflect its contribution given where its teammates stand; and (ii) variance reduction under the non-stationarity induced by concurrently updating policies. It is discarded at deployment, where each RSL executes its own actor using only its local observation.

*Training.* Advantages are estimated with generalized advantage estimation as:

$$\begin{cases} \delta_t = r_t + \gamma V(s_{t+1})(1 - \text{done}_t) - V(s_t) \\ A_t = \delta_t + \gamma\lambda(1 - \text{done}_t)A_{t+1} \end{cases} \tag{6}$$

All actors share the team advantage while being updated with the clipped PPO objective plus an entropy bonus as:

$$\begin{cases} L_{\text{actor}} = -\mathbb{E}[\min(\rho_i A_t,\ \text{clip}(\rho_i, 1-\epsilon, 1+\epsilon)A_t)] \\ \qquad\qquad - c_e\, \mathbb{E}[\mathcal{H}(\pi_i)] \\ L_{\text{critic}} = 1/2\, \mathbb{E}\big[(V(s_t) - \widehat{R}_t)^2\big] \end{cases} \tag{7}$$

where $\rho_i$ is the probability ratio between the new and old policies, the second term is an entropy bonus, and the value target is the advantage plus the current value estimate. We use discount factor $\gamma = 0.99$, GAE coefficient $\lambda = 0.95$, clipping parameter $\varepsilon = 0.2$, entropy coefficient 0.01, and learning rates $3\times10^{-4}$ (actor) and $10^{-3}$ (critic). Cooperation among agents emerges from four coupled mechanisms rather than explicit communication as follows: (i) the shared team reward, (ii) the centralized critic's conditioning on the joint configuration, (iii) the shared global encoding that persists at execution time, and (iv) the joint POG simulation through which any agent's adjustment changes every agent's next observation.

## 4. Experiments

We evaluate RSLaA on 10 real sites reconstructed from high-definition point clouds, covering diverse road geometries including urban arterials, intersections, and curves. These 10 experiment sites are disjoint from the 17 sites used for SM benchmarking. On these sites we construct 89 traffic scenarios in total, each carrying 6 hours of simulated traffic generated by the co-simulation platform described in the supplementary material. Scenarios span different traffic volumes, speeds, vehicle compositions, etc.

## 4.1. Experiment Setup

Each traffic scenario is evaluated under four deployment strategies: (i) Static, in which all RSL keep their initial poses throughout the 6-hour period, reproducing the conventional fixed-deployment paradigm; (ii) MPC, the receding-horizon surrogate optimization of Sec. 3.3; (iii) MARL-matched (MARL_m), in which the deployed MAPPO policy was trained with the same decision-window length as used at test time; and (iv) MARL-mismatched (MARL_mm), in which the policy was trained at a different window length, probing the cross-window transferability of the learned policies. Policies are pre-trained at all three window lengths (5, 15, 30 min). For each scenario we form 3 configuration units covering $w$ in {5, 15, 30} min, and each unit is evaluated under the three adaptive strategies, yielding 9 adaptive runs plus 1 static baseline, i.e., 10 runs per scenario and 890 runs in total. All methods in the same scenario share identical initial poses, sensor suites, traffic data, and evaluation parameters, so that any performance difference is attributable solely to the deployment strategy. Unless otherwise stated, experiments use $N$ = 2 heterogeneous LiDARs (HDL-32 + VLP-16), while the formulation and both methods support arbitrary $N$. Pose adjustment margins (position margin in {0.2, 0.5} m, angular margin in {5, 10, 15} deg) are assigned per configuration unit to probe the effect of the physically feasible adjustment range.

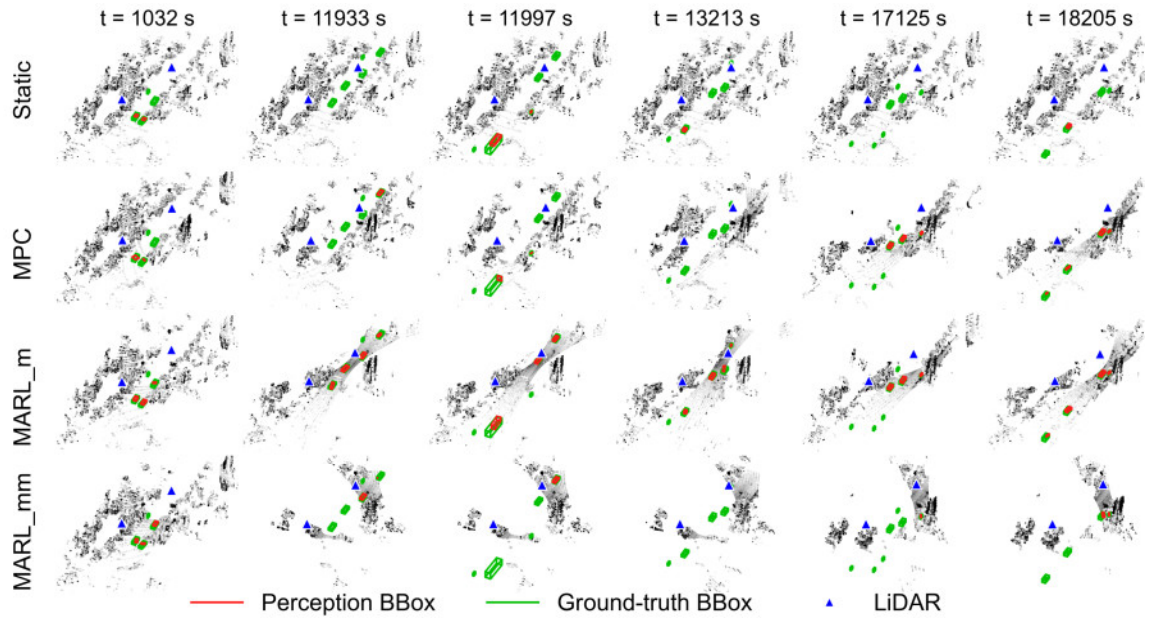

**Figure 6.** Point cloud sequence and perception labels corresponding to different method

To strengthen the learned policies, training is carried out across scenarios rather than within a single site. After completing training on one scenario, the policy weights are carried over and training continues on a new scenario, and in this study each policy is trained sequentially over three scenarios before deployment. This sequential cross-scenario training is applied independently to each of the three model window lengths.

## 4.2. Protocols and Evaluation Metrics

We first ran the MPC and the MARL policies on the 89 test scenarios, which produced 9×9×9+9×8=801 adaptive pose sequences: three MPC configurations, three MARL_m configurations, and three MARL_mm configurations. Together with the 89 fixed sequences of the static baseline this gives 890 sequences.

Executed pose sequences are replayed in the high-fidelity LiDAR simulator, as illustrated in Figure 6, and perception quality is measured against GT in a detector-agnostic manner. All metrics are computed at the object-frame level from per-target records. A valid detection (TP) requires a successful association together with a 3D IoU above 0.15, separating ‘detected but poorly localized’ from ‘not detected’. We report three indicators plus a composite score. IoU is the mean 3D IoU over all valid detections, measuring the quality of localization for targets that are actually perceived. Dev is the median bottom-center XY error over valid detections, chosen over the mean for robustness against degraded boxes. IvR is the invalid-box rate, i.e., the fraction of produced BBoxes that fail to qualify as valid detections.

## 4.3. Results and Comparisons

To remove the site-to-site variability, every frame is compared with the static result of the same scenario. Collapsing each run to one value and averaging the relative change over the 89 scenarios, as shown in Figure 7, gives a compact summary. IoU increases by 5.6%, 6.9% and 6.1% for MPC, MARL_m and MARL_mm, respectively; IvR decreases by 15.2%, 21.6% and 18.3%; and Dev decreases by 5.8%, 3.2% and 2.9%.

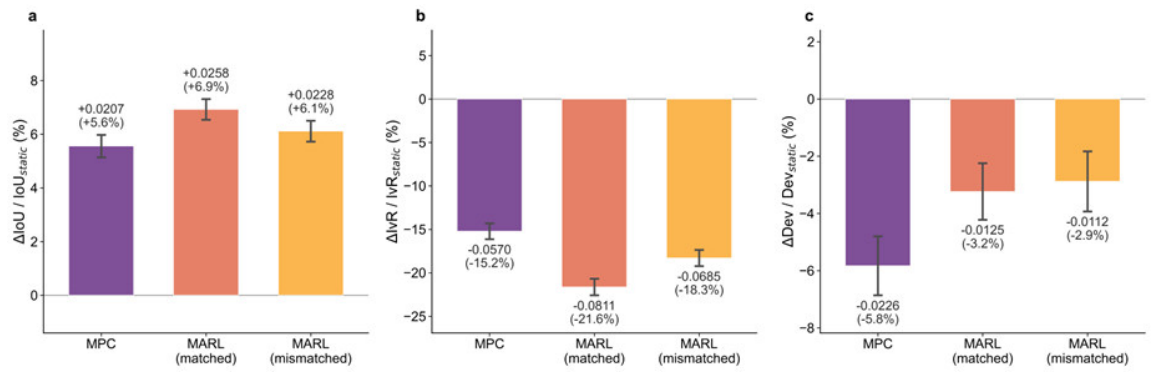

**Figure 7**. Relative change of the run-level metrics with respect to the static baseline: (a) IoU, (b) IvR, and (c) Dev. Error bars are bootstrap standard errors (400 resamples); the absolute differences are printed above and below the bars.

The sign of the change is consistent across the three controllers, so the improvement is not an artefact of a particular policy. In addition, the ranking is stable, MARL_m > MARL_mm > MPC on IoU and IvR, which indicates that a learned policy exploits the perception feedback more effectively than the model-predictive one, and that the memory horizon matters, but its effect is far smaller than the distance to the static baseline.

Table 1 resolves the same metrics per site. RSLaA improves IoU in 9 of the 10 sites, reduce IvR in all 10 sites, by 3.1 to 11.4 percentage points, and reduce Dev in 6 of the 10 sites, whereas sites 1, 2, 7 and 10 are worse by 0.01 to 0.03 m. The magnitude of the gain is scene-dependent. The largest improvements occur at sites 4 and 8 (+0.034 and +0.024 IoU, −7.7% and −5.2% IvR), which combine dense traffic with a narrow lane geometry, while site 5 already reaching the highest static IoU (0.404) with a small

**Table 1** Per-site perception metrics under the static baseline and the three adaptive configurations

| Site | IoU | | | | IvR (%) | | | | Dev (m) | | | |
|---|---|---|---|---|---|---|---|---|---|---|---|---|
| | **static** | **mpc** | **marl_m** | **marl_mm** | **static** | **mpc** | **marl_m** | **marl_mm** | **static** | **mpc** | **marl_m** | **marl_mm** |
| scene_1 | 0.328 | 0.360 | **0.362** | 0.361 | 36.89 | 30.57 | 28.35 | **28.27** | **0.398** | 0.418 | 0.436 | 0.428 |
| scene_2 | 0.361 | **0.385** | 0.381 | 0.380 | 33.49 | 28.95 | 27.41 | **26.94** | **0.343** | 0.363 | 0.394 | 0.409 |
| scene_3 | 0.352 | 0.385 | **0.391** | **0.391** | 38.20 | 32.27 | **28.03** | 29.25 | 0.421 | **0.372** | 0.383 | 0.377 |
| scene_4 | 0.381 | 0.405 | **0.415** | 0.412 | 32.07 | 26.50 | **24.38** | 26.00 | 0.373 | 0.318 | 0.330 | **0.327** |
| scene_5 | **0.404** | **0.404** | 0.403 | 0.403 | 39.82 | **34.50** | 36.44 | 36.71 | 0.307 | **0.282** | 0.288 | 0.287 |
| scene_6 | 0.351 | 0.382 | **0.391** | 0.387 | 43.63 | 38.09 | **33.50** | 34.38 | 0.445 | 0.396 | **0.388** | 0.404 |
| scene_7 | 0.366 | 0.389 | **0.393** | 0.382 | 44.12 | 36.80 | **32.73** | 35.07 | 0.427 | **0.382** | 0.383 | 0.416 |
| scene_8 | 0.414 | 0.434 | **0.438** | 0.434 | 27.90 | 23.08 | **22.70** | 23.23 | 0.313 | **0.287** | 0.290 | 0.299 |
| scene_9 | 0.396 | 0.409 | **0.419** | 0.413 | 33.63 | 29.06 | **25.33** | 26.19 | 0.347 | 0.330 | **0.328** | 0.343 |
| scene_10 | 0.400 | 0.410 | **0.413** | 0.402 | 31.10 | 28.45 | **25.17** | 27.14 | 0.372 | **0.342** | 0.370 | 0.375 |

deviation, which leaves little room for improvement.

The policies are also ranked differently from site to site, as presented in Table 1. MARL_m attains the best IoU in 6 of the 10 sites and the lowest IvR in 8 of them, and MARL_mm stays within ΔIoU of 0.011 of it everywhere, i.e. a memory horizon that does not match the evaluated traffic profile costs little once the policy is deployed. MPC is seldom the best (best IoU only at sites 1 and 2) but never degrades IoU or IvR with respect to static apart from a tie at site 5, which makes it the safer fallback when the learned policy operates out of distribution. Overall, Table 1 supports the central claim of this work that an adaptive configuration, independently of the controller, improves CoP compared with a fixed one, while showing that the size of the gain is governed by the scene rather than by the controller.

Figure 8 isolates the three design parameters by averaging $\Delta S_t$ over all runs that share the same value. Window length is the dominant factor (Figure 8a): $\Delta S_t$ increases monotonically with the window for all three policies, from about −0.02 at 5 min to +0.01 to +0.02 at 30 min. With a 5-min window the estimate is driven by too few frames and both MARL policies fall below the static baseline; at 15 min MARL_m already peaks, and at 30 min every policy is positive. Window length therefore trades responsiveness against estimate stability, and the useful range starts at 15 min.

The two margins are second-order (Figure 8a, 8b). Position margin has almost no effect. The three curves stay within ±0.01 and their confidence bands overlap over 0.2–0.5 m. Angle margin shows a small but consistent negative slope: for MARL_m, $\Delta S_t$ falls from +0.018 at 5° to about 0 at 15°, for MARL_mm, $\Delta S_t$ falls from +0.003 to −0.010, while MPC remains flat. Relaxing the margins admits configurations whose pose uncertainty exceeds the estimator error, so the extra freedom does not ensure a gain.

The practical implication is that the tuning space is effectively one-dimensional: use a window of 15 to 30 min and keep the margins tight (about 0.2 m and 5°). The remaining parameters, position margin in particular, can be left at their nominal values without loss, which simplifies

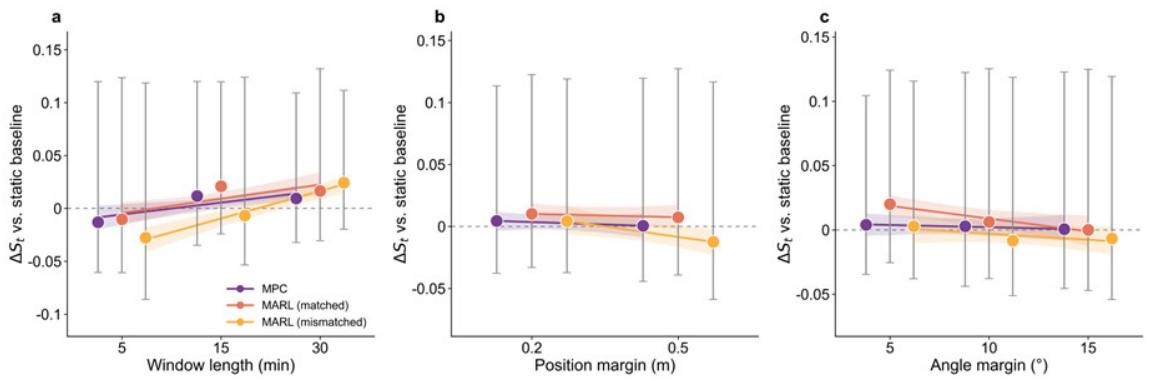

**Figure 8.** Effect of the design parameters on the score gain over the static baseline: (a) window length, (b) position margin, (c) angle margin. Markers are means over the runs sharing a parameter value, vertical bars are 95% confidence intervals and the dashed line marks the static level. Note that the composite score is $\boldsymbol{S_t} = \mathbf{0.4} \cdot \boldsymbol{IoU} + \mathbf{0.3} \cdot (\mathbf{1} - \boldsymbol{clip}(\boldsymbol{Dev}, \mathbf{0}, \mathbf{2})/\mathbf{2}) + \mathbf{0.3} \cdot (\mathbf{1} - \boldsymbol{IvR})$, with Dev clipped to [0, 2] m. Frames or runs containing no valid detection are filled with the worst-case values, so that missed detections are never masked by frame exclusion. The primary endpoint is the paired gain over the static baseline, $\boldsymbol{\Delta S_t} = \boldsymbol{S_{t(adaptive)}} - \boldsymbol{S_{t(static)}}$.

deployment and reduces the calibration effort.

We also conducted SHAP analysis to reveal the influence of different factor on $\Delta S_t$. The perception gain can be attributed to traffic, geometric, and configuration factors, as presented in the supplementary material.

## 5. Conclusion

This paper presented RSLaA, to our knowledge the first framework that treats roadside LiDARs as agents capable of adaptive pose reconfiguration in response to dynamic traffic. We formalized the problem around a causal, history-only SM, validated five candidate surrogates on a purpose-built co-simulation benchmark, and instantiated two adaptive configuration methods, receding-horizon optimization and multi-agent reinforcement learning upon it. Experiments over 10 sites and 89 six-hour scenarios demonstrate that SM-driven adaptive deployment outperforms static deployment with identical initial poses, with the gain varying systematically with traffic demand, road geometry, adjustment margins, and window length. Beyond roadside perception, the framework exemplifies a general principle: infrastructure sensors become more valuable when their data-acquisition geometry is treated as a decision variable rather than a fixed constant.

## 6. LLM Usage Statement

During the preparation of this work, the authors used LLM-based tools to assist with implementation and writing. Kimi-K3 was used for language polishing of the manuscript, DeepSeek V4.1 Flash was used to assist with the adjustment and optimization of code, and the TRAE agent was used to generate part of the result visualizations from the experimental data. The research conception, problem formulation, methodology design, and all conceptual figures were completed entirely by the authors without LLM involvement. The authors reviewed and verified all LLM-assisted content, including code and visualizations, and take full responsibility for the accuracy, originality, and integrity of this paper.

## Supplementary Material: Roadside LiDARs as Agents

# A. Surrogate Metrics

This section provides the complete definitions of the four surrogate metrics not detailed in the main text ($J_{\mathrm{EIG}}$, $J_{\mathrm{cov}}$, $J_{\mathrm{dyn}}$, $J_{\mathrm{traj}}$), together with the shared computational modules. The definition of $J_{\mathrm{mv}}$ is given in the main text; the remaining four including $J_{\mathrm{mv}}$ are defined here.

## A.1. Shared modules

All metrics operate on the probabilistic occupancy grid (POG), which stores a voxel occupancy probability $p_i$ and an observation mask $o_i$. Bird's-eye-view (BEV) column occupancy is obtained by pooling along the height axis as:

$$p_k^{\mathrm{col}} = \max_z p_{k,z} \tag{s1}$$

where $k$ indexes BEV columns and $z$ the height layers. The task region delimits the space worth observing and is constructed from the historical vehicle trajectories $\mathbf{\Gamma}$ by projecting them onto the BEV grid and dilating by the road half-width $r_{\mathrm{road}}$:

$$R = \mathrm{Dilate}(\mathrm{Proj}(\Gamma),\ \mathrm{round}(r_{\mathrm{road}}/\Delta)) \tag{s2}$$

where voxel size Δ = 0.4 m and $r_{\mathrm{road}}$ = 5 m. The column-level soft visibility of column $k$ for LiDAR $n$ under a candidate pose combines a distance decay with a line-of-sight transmittance estimated over the historical column occupancy:

$$q_{n,k}^{col} = \frac{1}{1+(r_k/\rho_n)^2} \prod_{s \in S_k} (1 - \alpha\, p^{col}(s)) \tag{s3}$$

where $r_k$ is the horizontal distance from the LiDAR to the column center, $\rho_n$ the reference range of the sensor model, $S_k$ the set of $n_{\mathrm{ray}}$ samples along the 2D sight line (excluding both endpoints), and α = 0.9 the occlusion attenuation coefficient. Voxel-level visibility augments the column value with a pitch field-of-view mask $M_{i,n}$ and a 3D-to-2D decay ratio:

$$q_{i,n} = q_{n,k(i)}^{\mathrm{col}} \cdot M_{i,n} \cdot (r_i^{(2)}/r_i^{(3)})^2 \tag{s4}$$

The joint visibility of $N$ LiDARs assumes observation independence:

$$q_i = 1 - \prod_{n=1}^{N} (1 - q_{i,n}) \tag{s5}$$

The uncertainty of an occupancy state is measured by the Bernoulli entropy

$$H(p_i) = -p_i \ln p_i - (1-p_i)\ln(1-p_i) \tag{s6}$$

The voxel task weight $w_i$ equals 1 for voxels lying in a task-region column and within the field of view and range of at least one RSL, and 0 otherwise, so that the metric denominators count exactly the voxels that should be observable.

## A.2. The four remaining surrogates

**Expected information gain** $J_{\mathrm{EIG}}$ prioritizes coverage of voxels whose occupancy state is still uncertain, approximating the expected entropy reduction upon observation:

$$J_{\text{EIG}} = \frac{\sum_i w_i \; q_i \; H(p_i)}{\sum_i w_i} \tag{s7}$$

**Task-weighted coverage** $J_{\text{cov}}$ is the fraction of the task region that is actually visible:

$$J_{\text{cov}} = \frac{\sum_i w_i \; q_i}{\sum_i w_i} \tag{s9}$$

**Corridor continuity** $J_{\text{traj}}$ evaluates whether visibility is continuous along predicted traffic corridors. Historical passable columns are grouped into corridors by principal-component decomposition (8 lateral bands, 5 m segments); each corridor $c$ receives a confidence $a_c$ (twice the mean occupancy, clipped to [0, 1]) and a score:

$$s_c = \frac{1}{N_c} \sum_{b=1}^{N_c} Q_b \; - \; \mu \, L_c, J_{\text{traj}} = \frac{\sum_c a_c \; s_c}{\sum_c a_c} \tag{s10}$$

where $Q_b$ is the mean joint column visibility in segment $b$, $N_c$ the number of segments, $L_c$ the number of consecutive low-visibility runs (at least 2 consecutive segments with $Q_b$ below $q_{\text{low}}$ = 0.3), and $\mu$ = 0.5 the continuity penalty. The metric may take negative values when long corridors are mostly invisible, which is by design.

**Dynamic weight** $J_{\text{dyn}}$**.** The dynamic weight used by $J_{\text{dyn}}$ is

$$d_i = \frac{\sum_r \eta^{R-1-r} \; o_i^r o_i^{r-1} \; \text{JSD}_i^{(r-1,r)}}{\sum_r \eta^{R-1-r} \; o_i^r o_i^{r-1}} \tag{s11}$$

where temporal decay $\eta$ = 0.8 in this case and $d_i$ degenerates to the Bernoulli entropy when fewer than two historical windows are available.

**Direction conventions and degenerate handling.** The usable direction of each surrogate is fixed by its measured correlation with perception performance in the benchmark rather than by intuition: $J_{\text{dyn}}$ and $J_{\text{traj}}$ are minimized, while $J_{\text{cov}}$ , $J_{\text{EIG}}$ and $J_{\text{mv}}$ are maximized. When a metric is not applicable at a decision point, no historically dynamic region for $J_{\text{dyn}}$, or no reliable corridor for $J_{\text{traj}}$, it returns a null marker instead of a numerical value and is excluded pairwise from the correlation analysis. If the trajectory-based task region is empty, it falls back to a dilation of historically active columns, then to the historically observable region, and finally to the full grid. There are two possible properties that make $J_{\text{mv}}$ suitable as a decision core in this study. $J_{\text{mv}}$ is computable entirely from history available at decision time, and it targets precisely the aspect of multi-LiDAR cooperation that single-sensor coverage cannot capture, which is the triangulation geometry that governs the quality of 3D perception. The benchmark further confirms this intuition: among all five surrogates, $J_{\text{mv}}$ achieves the strongest correlation with the composite perception score, with consistent signs across the window-length and margin strata. Importantly, the usable direction of every surrogate is determined empirically by its measured correlation with perception performance rather than by design intuition; for $J_{\text{mv}}$ the benchmark yields a positive correlation with the composite perception score, and accordingly both adaptive methods maximize $J_{\text{mv}}$, it enters the MPC objective and the MARL team reward $r_t$ directly.

## B. Benchmark Construction Details

On each site, simulated vehicle traffic from 25 trajectories (21600 frames per trajectory) is injected into the static map, and the measurement period is tiled into gap-free rolling windows of 5, 15, and 30 min (300/900/1800 frames at 1 Hz), yielding 2,522 windows in total, of which 2,449 involve a configuration decision. At each decision window, six road-side LiDAR configurations (2-3 sensors drawn from VLP-16, HDL-32E, and OS1-64, each perturbed within +-0.6 m in position and +-10 deg in roll/pitch around the previously selected pose) are proposed. Each candidate is scored by the five surrogate metrics using only the preceding windows' occupancy history, and is then re-simulated at high fidelity to obtain three perception labels: IoU, mean position deviation (Dev, m), and sequence completeness (Comp), as visualized in Figure S1. This produces 15 surrogate-perception metric pairs per candidate, i.e., 14,694 labelled candidate configurations and 220,410 pairs, of which 211,938 remain valid after discarding the degenerate cases. The data structure is presented in Figure S2.

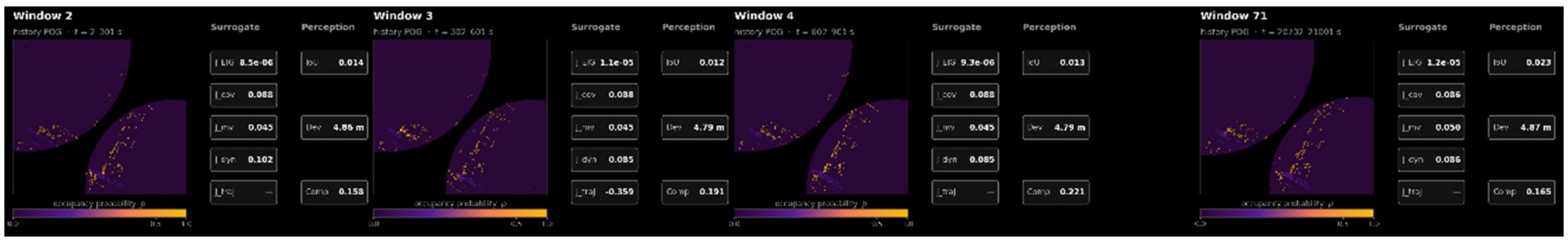

**Figure S1.** Examples of POG-based SM and Paired Perception Labels

During the creation, the first window establishes the initial POG from randomized pole-mounted poses and produces no decision record. From the second window onward, six candidate adjustment plans are sampled within the margins around the previously selected configuration; the five surrogates are computed from the historical POG alone, and each candidate is then executed in high-fidelity simulation for the next window to obtain IoU, Dev, and Comp labels via point-based bounding-box fitting. The best-scoring candidate is executed, the POG rebuilt, and the process rolls forward. Traffic statistics (mean speed, volume) of every window are logged for stratified analysis.

For each surrogate we compute Pearson and Spearman correlations against the composite score and against each individual label, both over the full record set and stratified by window length and margin; significance and the effective sample size after pairwise null deletion are reported. Selection validity is assessed by the true-score rank distribution of the surrogate-preferred candidate among the six candidates of each window.

```
bench_n/
├── scene_001/
│   ├── scene_info.json
│   └── traffic_01/
│       ├── traffic_info.json
│       ├── t_05/
│       │   ├── config.json
│       │   ├── lidar_para_001_t_05_000001.json
│       │   ├── pog_001_t_05_000001.npz
│       │   ├── traffic_001_t_05_000001.txt
│       │   ├── data_001_t_05_000002.json      # only k ≥ 2
│       │   └── viz_001_t_05_000002.png        # for --viz
│       ├── t_15/
│       └── t_30/
├── scene_002/
└── ...
```

**Figure S2.** Data structure of benchmark

## C. Experiment Sites

The benchmark is built on 17 real-world point-cloud sites: 16 segments from five KITTI-360 drives and one mobile-mapping site from WHU-Urban3D, as shown in Figure S3(a). The 10 experiment sites, as displayed in Figure S3(b), are reconstructed from real roadside point clouds and were selected to cover the geometric diversity of typical roadside deployments: straight urban arterials, signalized and unsignalized intersections, and curved road segments with roadside obstacles. These sites are disjoint from the benchmark sites of Sec. B, so that surrogate selection and method evaluation do not share data.

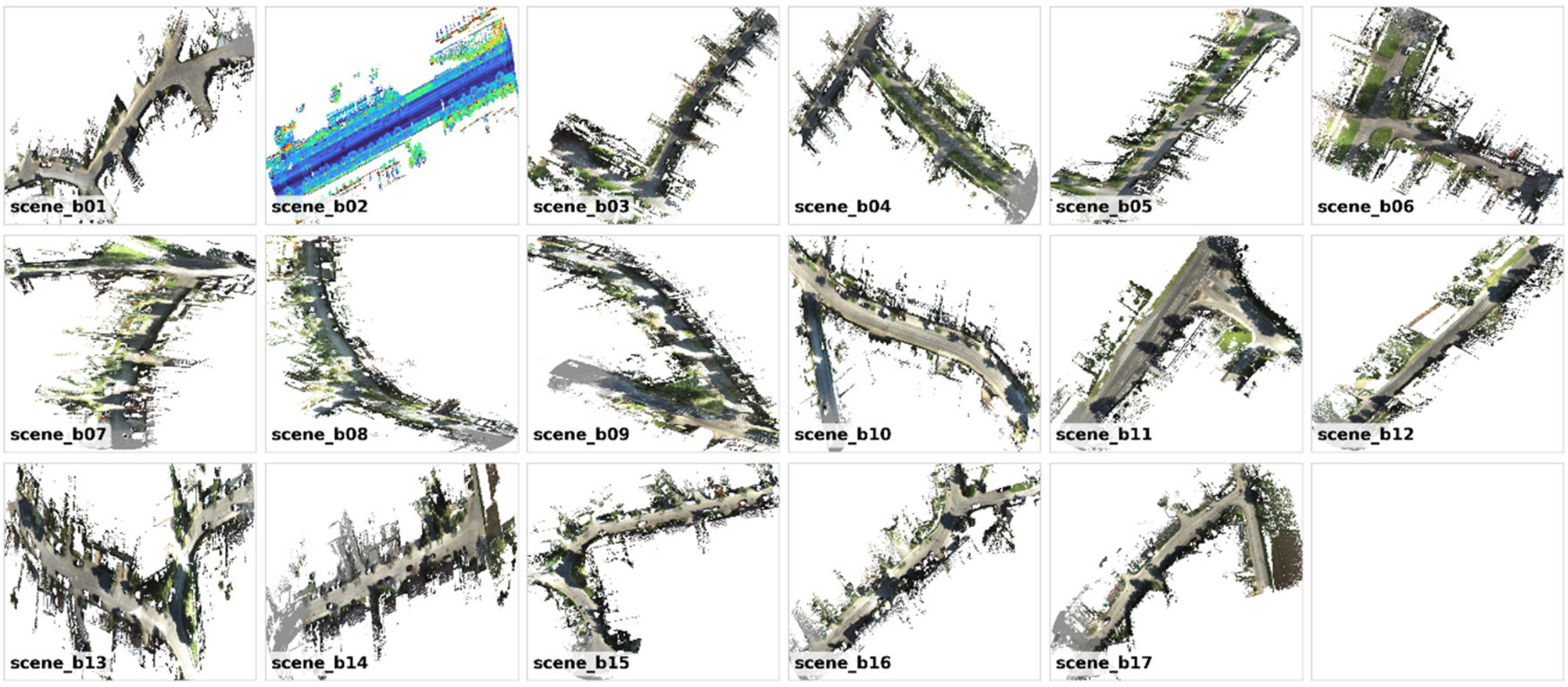

(a)

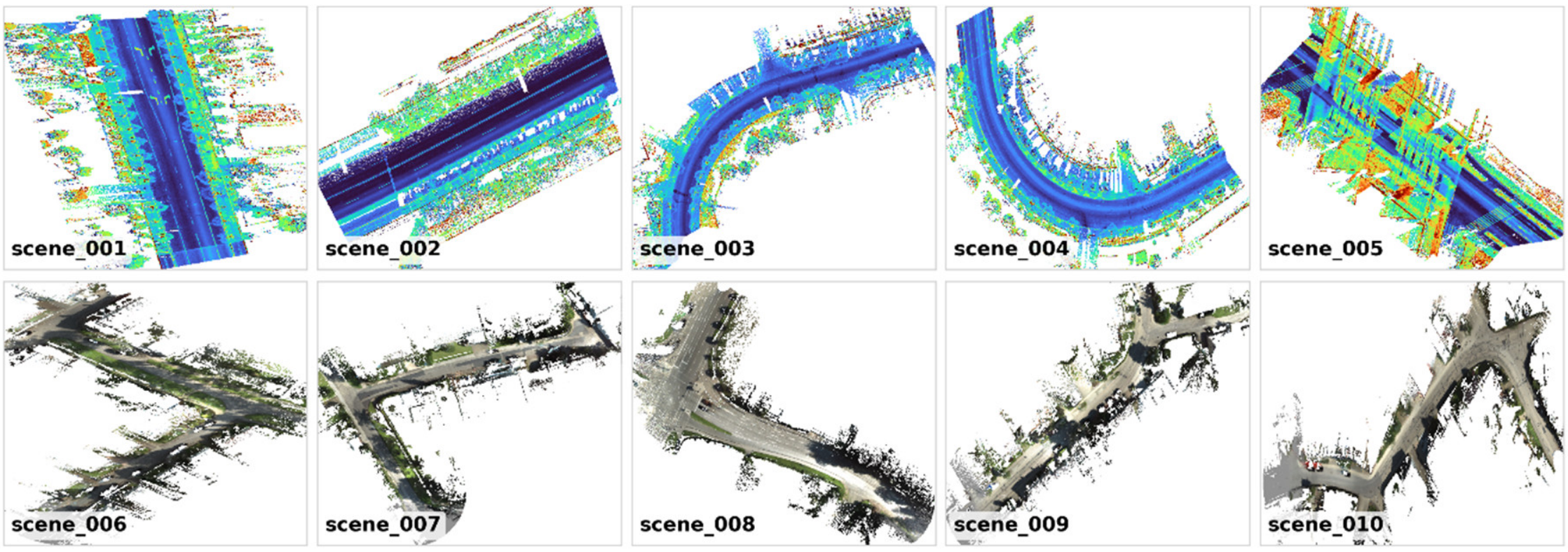

(b)

**Figure S3.** Visualization of Sites Used for Benchmark Creation and Experiment (a) Benchmark creation, and (b) Experiment

## D. Distribution of Perception Gains

To remove the site-to-site variability, every frame is compared with the static result of the same scenario, so that all points in Figure S4 are paired differences. For all three policies the distribution of ΔIoU is centered above zero (median ≈ +0.01 to +0.02) and that of ΔIvR below zero (≈ −0.04 to −0.06): in the majority of frames the adaptive configurations accept more valid boxes and reject fewer. The ΔDev distribution is much narrower and slightly positive (≈ +0.03 m), i.e. the median frame is marginally worse in centroid accuracy; at the pooled object level the same comparison gives a 3 to 6% lower deviation, so this effect is confined to individual frames and does not accumulate over a run.

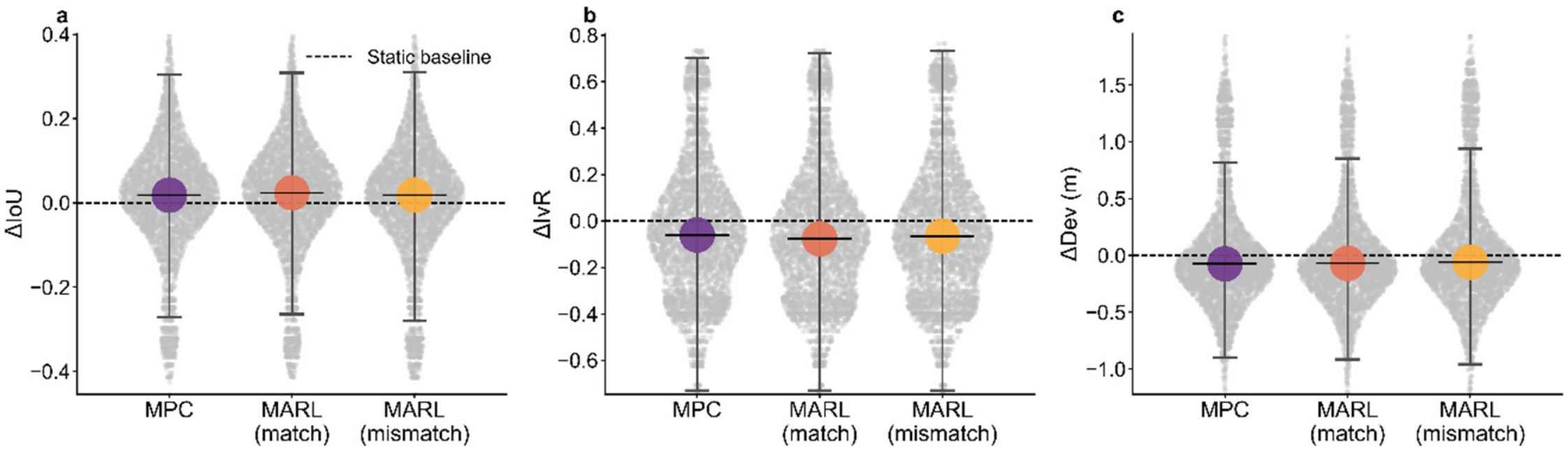

**Figure S4**. Frame-wise metric differences with respect to the static baseline of the same scenario. Grey points are individual frames (subsampled for readability), colored circles the medians and vertical lines the 1.5 IQR whiskers; the dashed line marks the static level.

## E. What Drives the Gain?

Figure S5 attributes the per-run $\Delta S_t$ to scene covariates and design parameters with a gradient-boosted model and TreeSHAP. The ranking is dominated by the scene: mean speed (mean |SHAP| = 0.0111, 31%), window length (0.0092, 26%), lane width (0.0057, 16%), angle margin (0.0051, 14%) and traffic volume (0.0051, 14%), followed by the car/truck ratio (0.0043). Layout descriptors such as obstacle height, longitudinal range, road width, number of intersections or lanes and curvature contribute little, and position margin is the least important feature of all (0.0001).

The dependence plots give the sign of these effects. The gain decreases with mean speed. Positive SHAP values below about 24 km/h and negative values above about 25 km/h, and increases with traffic volume (negative near 1,000 veh/h, positive near 2,600 veh/h) and with the car/truck ratio: adaptive reconfiguration pays off when many small, densely packed objects have to be resolved. Lane width acts in the same direction, the contribution being positive at about 2.5 m and negative at 3.2 to 3.5 m, so narrow corridors benefit from re-optimizing the sensor geometry whereas wide roads do not. Among the design parameters, window length is the only one with a monotonic positive effect, and angle margin reduces the response, consistent with Figure 8 in the main text.

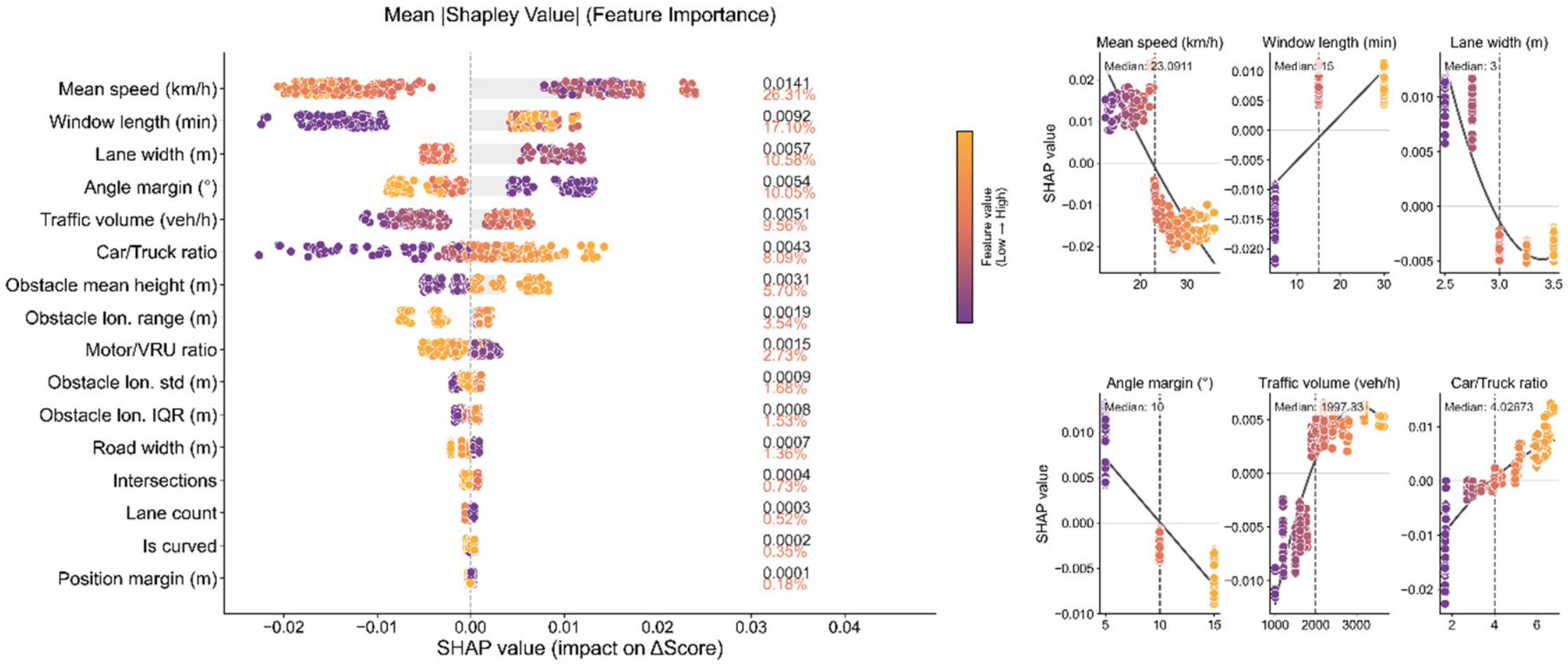

**Figure S5**. SHAP attribution of $\Delta S_t$. Left: mean |SHAP| per feature, with the value printed beside each bar and the relative contribution in brackets. Right: dependence of the SHAP value on the six most important features; the dashed lines mark the median of each feature and colours encode the feature value.

Based on Figure S5 and Figure 8 in the main text, the benefit of an adaptive configuration is predictable from the context, and once the context is favorable the controller choice and the margin settings are second-order. This motivates a context-aware deployment strategy: predict whether a site will benefit (speed, volume and lane geometry), and then use a 15 to 30 min window with tight margins instead of tuning one global configuration for every site.

## F. Key Hyperparameters

**Table S1. MPC configuration**

| Parameter | Value |
|---|---|
| Decision-window length | 5/15/30 min |
| Position / angular margin | 0.2 or 0.5 m/5, 10, 15 deg |
| Optimizer | gradient-free candidate search + Nelder–Mead refinement |
| Random candidates per window | 64 (+ 3 deterministic seeds) |
| Refinement evaluation budget | ≤ 120 |
| ROI column subsampling | ≤ 12,000 columns, fixed seed |
| Line-of-sight samples (optimization / reporting) | 24 / 48 |
| Occlusion attenuation $\alpha$ | 0.9 |
| Temporal decay $\eta$ | 0.8 |
| History windows retained | 5 |
| Voxel size / road half-width / z range | 0.4 m / 5 m / [−0.5, 6.0] m |

**Table S2. MARL (MAPPO) configuration**

| Parameter | Value |
|---|---|
| Observation / action / state dims | 127 (6 + 49 + 72) / 6 / 6N + 72 |
| Actor network | MLP 128 × 2, tanh; diagonal Gaussian; deterministic mean at deployment |
| Critic network | MLP 256 × 2, tanh; centralized (training only) |
| Local BEV / global encoding | 7 × 7 × 0.4 m / 8 × 8 × 8 m + 8 statistics |
| Training position / angular margin | 0.5 m / 15 deg |
| Discount $\gamma$ / GAE $\lambda$ | 0.99 / 0.95 |
| PPO clip $\varepsilon$ / epochs / batch | 0.2 / 4 / 64 |
| Learning rates (actor / critic) | $3\times10^{-4}$ / $1\times10^{-3}$ |
| Entropy coefficient / gradient clip | 0.01 / 0.5 |
| Training episodes per window length | 100 |
| Reward | r = −J_dyn (n_ray = 24), shared team scalar |
| Initial-pose randomization | enabled during training, disabled at evaluation |

**Table S3. Surrogate-metric constants**

| Parameter | Value |
|---|---|
| Voxel size $\Delta$ | 0.4 m |

| Parameter | Value |
|---|---|
| Road half-width $r_{\text{road}}$ | 5 m |
| Line-of-sight samples (benchmark) | 48 |
| Occlusion attenuation $\alpha$ | 0.9 |
| Temporal decay $\eta$ | 0.8 |
| Multi-view weight $\lambda$ ($J_{\text{mv}}$) | 0.5 |
| Continuity penalty $\mu$ / low-visibility threshold $q_{\text{low}}$ | 0.5 / 0.3 |
| Corridor bands / segment length | 8 / 5 m |

## G. Co-simulation Tool

The simulation environment is built upon real-world 3D road scene data from WHU-Urban3D[1] and KITTI-360[2], which provide surveyed roadside geometry (road surfaces, buildings, poles, and roadside obstacles) as the physical substrate of all experiments.

High-precision road network maps are manually drawn over the real scene geometry, and the microscopic traffic simulator (PTV-Vissim) generates vehicle trajectories under manually configured traffic parameters (demand levels, speed distributions, vehicle-type compositions, and directional splits), yielding 89 diverse and 6-hour traffic scenarios for the main experiments and the scenario sets for the benchmark. The general workflow is shown in Figure S6.

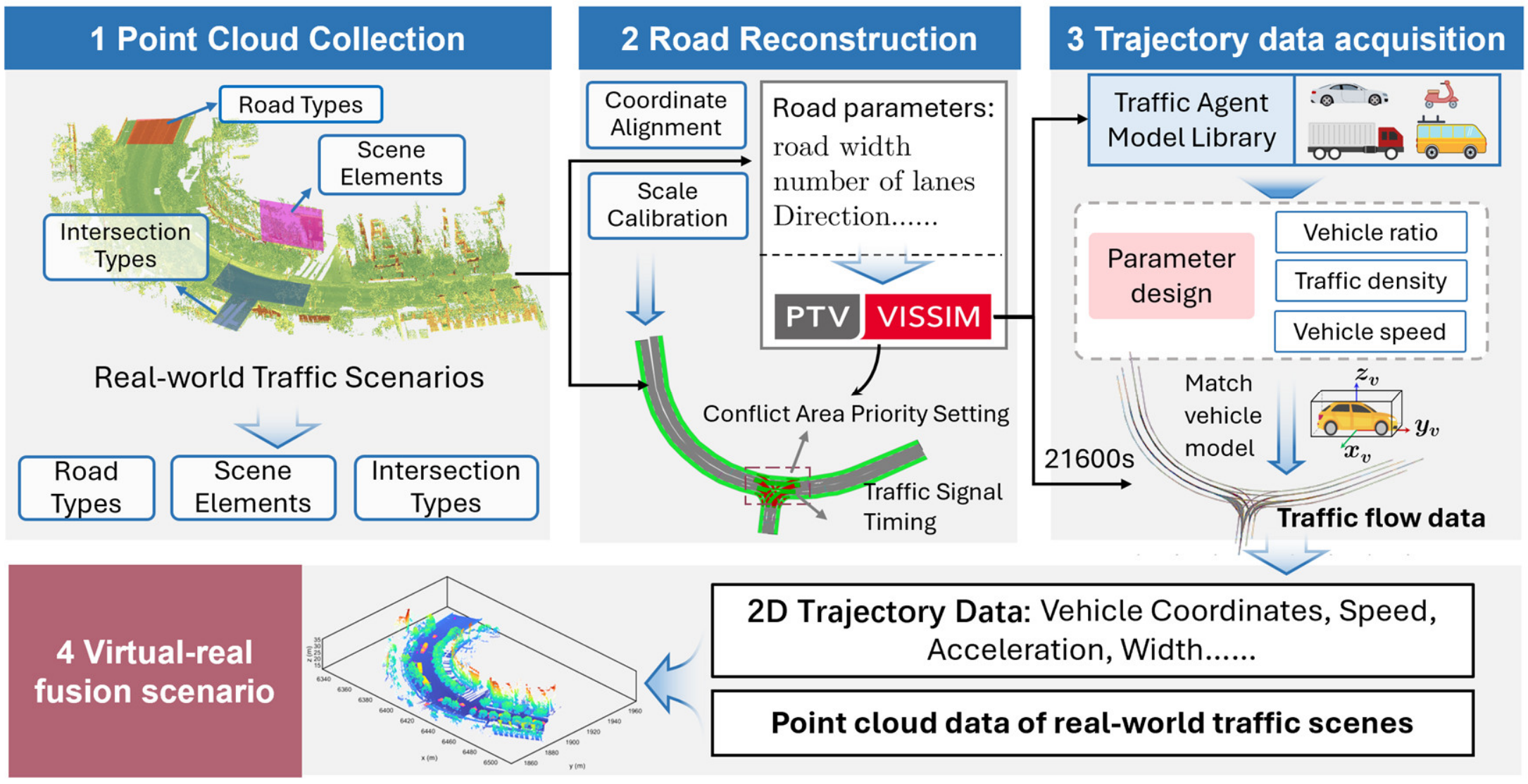

**Figure S6.** Workflow of Scenario Creation

A high-fidelity LiDAR simulator[3] renders physically realistic point clouds for heterogeneous sensor models (VLP-16, HDL-32, OS1-64, RS-128) under arbitrary 6-DoF poses, modeling per-beam ray casting over the scene mesh and moving vehicle mesh models, as illustrated in Figure S7. Executed pose sequences are replayed window by window; perception labels are obtained in a detector-agnostic manner by fitting per-vehicle bounding boxes from the simulated points on each target and comparing them with ground-truth envelopes (IoU, Dev, Comp in the benchmark; IoU, Dev, MR in the main experiments). The validity of this simulation pipeline for roadside perception studies has been tested and verified in prior studies[4].

---

[1] Han, X., Liu, C., Zhou, Y., Tan, K., Dong, Z., Yang, B. (2024). WHU-Urban3D: An urban scene LiDAR point cloud dataset for semantic instance segmentation. ISPRS Journal of Photogrammetry and Remote Sensing, 209, 500–513.

[2] Liao, Y., Xie, J., Geiger, A. (2022). Kitti-360: A novel dataset and benchmarks for urban scene understanding in 2d and 3d. IEEE Transactions on Pattern Analysis and Machine Intelligence, 45(3), 3292-3310.

[3] Cai, X., Jiang, W., Xu, R., Zhao, W., Ma, J., Liu, S., Li, Y. (2023, May). Analyzing infrastructure lidar placement with realistic lidar simulation library. In 2023 IEEE International Conference on Robotics and Automation (ICRA) (pp. 5581-5587). IEEE.

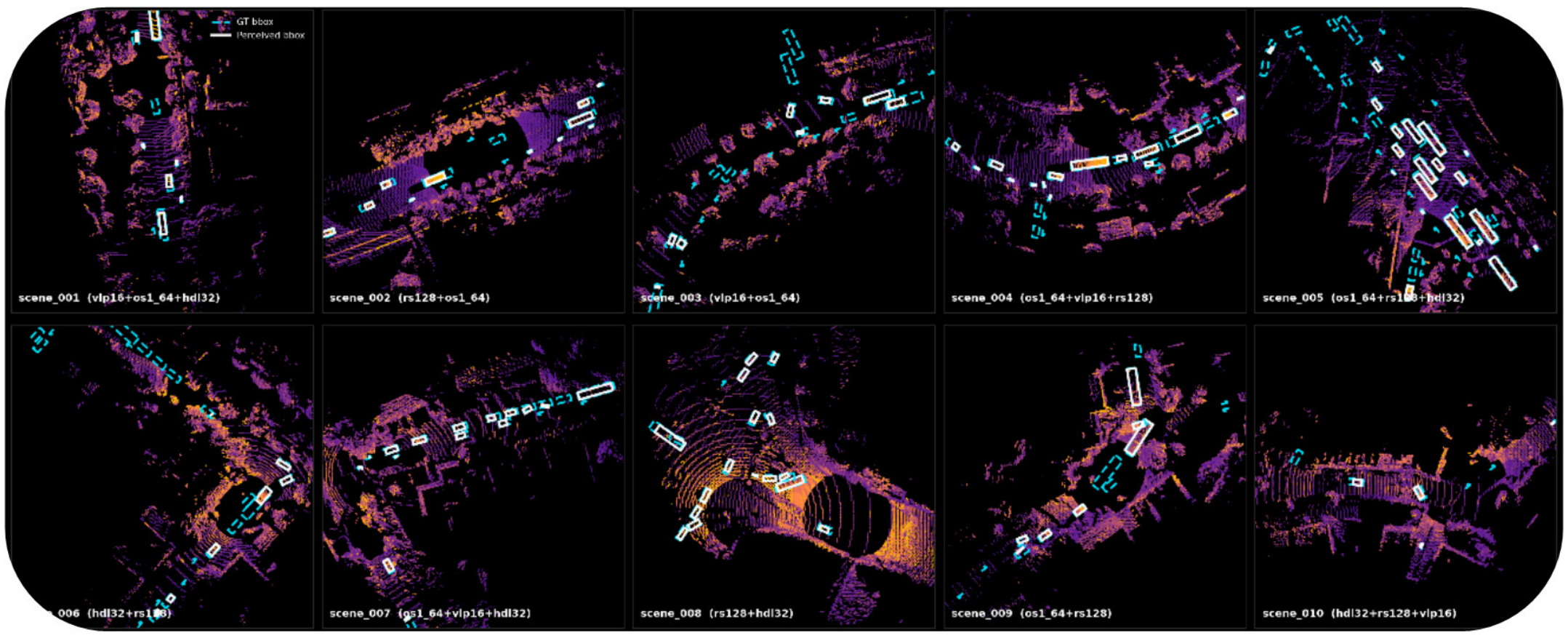

**Figure S7.** Lidar Simulation in Diverse Scenarios

## Implications and Limitations

The demonstrated benefit of adaptive configuration carries three practical implications. First, it offers a path to increase the effective value of costly roadside LiDAR deployments without additional sensors: with affordable multi-degree-of-freedom mounts, part of the performance gap between sparse and dense deployments can be recovered in software. Second, the surrogate-metric paradigm decouples data-acquisition quality from any specific perception model, so the framework remains compatible with future detection and fusion algorithms; the surrogate, benchmark, and adaptive methods constitute three independent entry points for follow-up research. Third, the same decision formulation, causal self-evaluation plus horizon-limited reconfiguration, extends naturally to other infrastructure sensors and to mobile or re-deployable sensing platforms.

Several limitations qualify our conclusions. (1) Simulation-to-real gap. All results are obtained in co-simulation; although the simulator is validated against real surveys, mechanical actuation limits, calibration drift after pose changes, sensor degradation, and communication latency in field deployments are not modeled, so the reported gains are upper estimates of deployable benefit. (2) Static history without prediction. The surrogate is computed from historical POGs and assumes short-horizon persistence of traffic structure; abrupt demand shifts or incidents cannot be anticipated, and the predictive horizon of the MPC branch is a single window. (3) Fidelity gap of the fast objective. Optimization uses a column-level approximation while reporting uses exact voxel-level metrics; the ranking consistency between the two, though empirically supported, is an approximation. (4) Scalability and long-horizon behavior. Main experiments use two LiDARs per scene; the random-search component faces the curse of dimensionality as the joint pose space grows to 6N dimensions, and box constraints relative to the previous window permit unbounded cumulative drift over long deployments, which hardware limits would need to bound. (5) Metric and evaluation scope. The invalid-box rate (MR) is a precision-like rather than a recall measure, and the SHAP analysis reveals associations rather than causal effects. (6) Learned-policy generality. MARL policies are trained sequentially over a limited number of scenarios; their transfer to unseen road geometries and traffic regimes, despite the rotation- and translation-invariant observation design, remains to be established in broader deployments.

---

[4] Ma, Y., Zheng, Y. B., Wang, S. Y., Wong, Y. D., & Easa, S. M. (2024). Virtual-real-fusion simulation framework for evaluating and optimizing small-spatial-scale placement of cooperative roadside sensing units. Computer-Aided Civil and Infrastructure Engineering, 39(5), 707-730.